\documentclass[conference]{IEEEtran}
\IEEEoverridecommandlockouts
\usepackage{cite}
\usepackage{amsmath,amssymb,amsfonts}
\usepackage{algorithmic}
\usepackage{graphicx}
\usepackage{textcomp}
\usepackage{xcolor}

\usepackage{listings}
\usepackage{xcolor}
\usepackage{hyperref}
\usepackage[utf8]{inputenc}
\usepackage{float}
\usepackage{multirow}
\usepackage{setspace}
\usepackage{longtable}
\usepackage{xltabular}
\usepackage{subcaption}
\usepackage{grffile}
\usepackage{booktabs}
\usepackage{makecell}

\makeatletter
\newcommand{\linebreakand}{%
  \end{@IEEEauthorhalign}
  \hfill\par%
  \mbox{}\hfill\begin{@IEEEauthorhalign}
}
\makeatother

\def\BibTeX{{\rm B\kern-.05em{\sc i\kern-.025em b}\kern-.08em
    T\kern-.1667em\lower.7ex\hbox{E}\kern-.125emX}}
\begin{document}
\setlength{\abovedisplayskip}{3pt}
\setlength{\belowdisplayskip}{3pt}

\title{MovieTeller: Tool-augmented Movie Synopsis with ID Consistent Progressive Abstraction\thanks{This work was supported by the Fundamental Research Funds for the Central Universities (No. 226-2025-00167), the National Natural Science Foundation of China (No. 62576308), and the Research Fund for International Scientists of National Natural Science Foundation of China (No. 72350710798). \copyright 2026 IEEE.}
}

% \textit{dept. name of organization (of Aff.)} \\

\author{\IEEEauthorblockN{Yizhi Li}
\IEEEauthorblockA{
\textit{Zhejiang University}\\
Hangzhou, China \\
yizhi.24@intl.zju.edu.cn}
\and
\IEEEauthorblockN{Xiaohan Chen}
\IEEEauthorblockA{
\textit{Zhejiang University}\\
Hangzhou, China \\
xiaohan.25@intl.zju.edu.cn}
\and
\IEEEauthorblockN{Miao Jiang}
\IEEEauthorblockA{
\textit{China Media Group}\\
Beijing, China \\
mylove\_513@hotmail.com}
\linebreakand
\IEEEauthorblockN{Wentao Tang}
\IEEEauthorblockA{
\textit{Watch AI Group}\\
Beijing, China \\
tangwent@sina.com}
\and
\IEEEauthorblockN{Gaoang Wang$^{\ast}$}
\IEEEauthorblockA{
\textit{Zhejiang University}\\
Hangzhou, China\\
gaoangwang@intl.zju.edu.cn\thanks{$^{\ast}$Corresponding author.}}
}

\maketitle

\begin{abstract}
With the explosive growth of digital entertainment, automated video summarization has become indispensable for applications such as content indexing, personalized recommendation, and efficient media archiving.
Automatic synopsis generation for long-form videos, such as movies and TV series, presents a significant challenge for existing Vision-Language Models (VLMs). While proficient at single-image captioning, these general-purpose models often exhibit critical failures in long-duration contexts, primarily a lack of ID-consistent character identification and a fractured narrative coherence. To overcome these limitations, we propose \textbf{MovieTeller}, a novel framework for generating movie synopses via tool-augmented progressive abstraction. Our core contribution is a training-free, tool-augmented, fact-grounded generation process. Instead of requiring costly model fine-tuning, our framework directly leverages off-the-shelf models in a plug-and-play manner. We first invoke a specialized face recognition model as an external ``tool" to establish Factual Groundings—precise character identities and their corresponding bounding boxes. These groundings are then injected into the prompt to steer the VLM's reasoning, ensuring the generated scene descriptions are anchored to verifiable facts. Furthermore, our progressive abstraction pipeline decomposes the summarization of a full-length movie into a multi-stage process, effectively mitigating the context length limitations of current VLMs. Experiments  demonstrate that our approach yields significant improvements in factual accuracy, character consistency, and overall narrative coherence compared to end-to-end baselines.
\end{abstract}

\begin{IEEEkeywords}
Video Summarization, Vision-Language Models, Tool-Augmented LLMs, Training-Free, Progressive Abstraction
\end{IEEEkeywords}

\section{Introduction}
The proliferation of digital video content, particularly long-form narratives such as movies and television series, has created an unprecedented demand. Generating a concise, coherent, and factually accurate synopsis of a full-length movie is a critical task with applications ranging from content recommendation and archival to aiding viewers with accessibility needs. However, this task poses a formidable challenge for current artificial intelligence systems. Despite the remarkable progress of Vision-Language Models (VLMs) in interpreting static images, their direct application to long video understanding reveals significant limitations. As shown in \textbf{Figure \ref{face_ground}}, we identify two primary bottlenecks that hinder the generation of high-quality movie synopses:

\begin{figure}[t]
    \centering
    \includegraphics[width=0.95\linewidth]{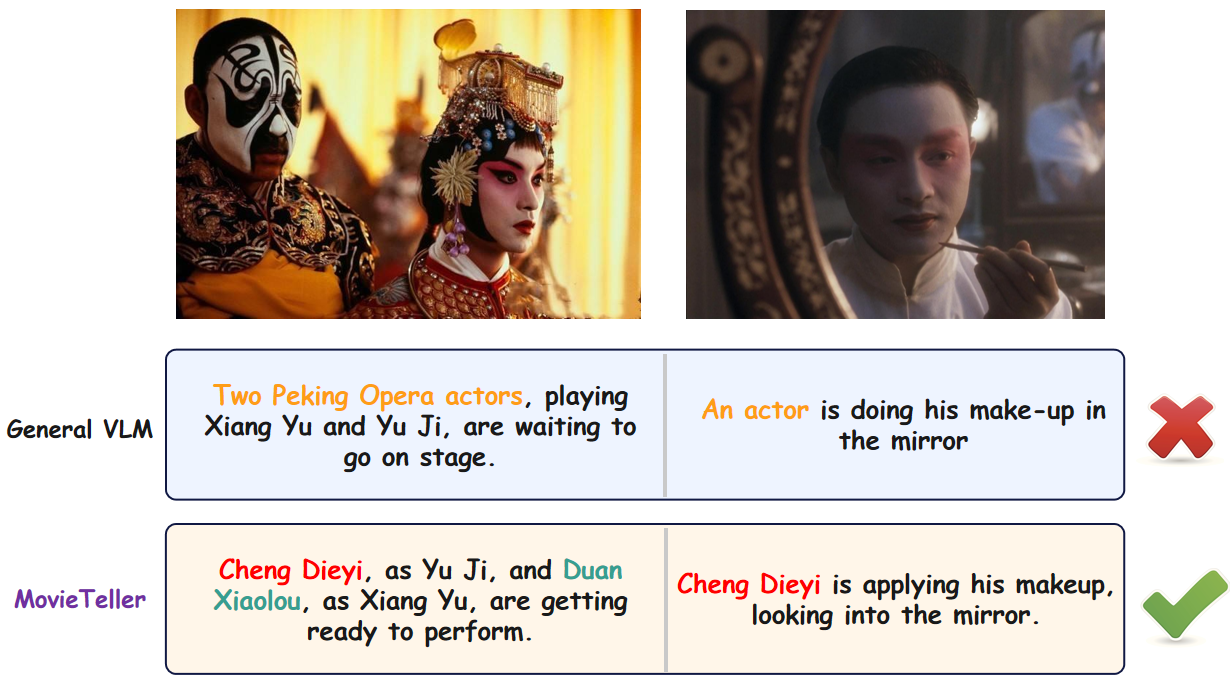}
    \caption{The advantage of our MovieTeller over general VLM. General VLM only plainly state the content of the frames, lacking ID consistency and narrative coherence. MovieTeller can ensure the accurate identification of characters, thereby guaranteeing the integrity and continuity of movie synopses.}
    \label{face_ground}
    \vspace{-5pt}
\end{figure}

\begin{itemize}
\item \textbf{Lack of ID-Consistent Character Identification:} General-purpose VLMs often struggle to recognize and consistently track specific characters throughout a lengthy narrative. They may describe a key protagonist as ``a man" in one scene and ``a person" in another, failing to bind the visual representation to a consistent identity. 

\item \textbf{Fractured Narrative Coherence:} The quadratic complexity of the Transformer's self-attention mechanism 
renders full-movie frame sequence processing computationally prohibitive.

Consequently, methods relying on uniform frame sampling or simple concatenation frequently disrupt narrative coherence, yielding fragmented summaries rather than unified stories.
\end{itemize}

To address these challenges, we introduce MovieTeller, a novel framework for generating movie synopses via tool-augmented progressive abstraction. Instead of attempting to build a single, monolithic model to solve all sub-problems, MovieTeller orchestrates a modular, training-free pipeline. The core idea is to augment a powerful, generalist VLM with an external, specialist ``tool" for a task it performs poorly on—face recognition. Specifically, we first invoke a high-accuracy face recognition model to establish factual foundations: the precise identities and locations of characters within key scenes. This factual information is then used to steer the VLM, compelling it to generate descriptions that are anchored to the ground truth of ``who is who" and ``who is where."
Furthermore, to tackle the long-context problem, MovieTeller implements a progressive abstraction pipeline. This process first condenses low-level scene descriptions into more abstract, information-dense chapter-level summaries. These chapter summaries are then integrated to produce the final, holistic movie synopsis. This multi-stage approach not only makes the computation tractable but also mirrors the human cognitive process of understanding complex narratives.

Our main contributions can be summarized as follows:

\begin{itemize}
\item We propose MovieTeller, a novel, training-free framework that generates ID-consistent and narratively coherent synopses for long videos. It effectively addresses the critical limitations of existing VLMs in handling character identification and long-range dependencies.

\item We introduce a plug-and-play architecture that operationalizes two key principles: tool-augmentation, where a specialist face recognition tool provides factual groundings to a generalist VLM; and progressive abstraction, which systematically condenses information from scenes to chapters and finally to a full synopsis. 

\item We validate our approach through extensive experiments and ablation studies on a diverse collection of 100 full-length films, totaling over 10,000 minutes of video content. The results demonstrate that MovieTeller achieves state-of-the-art performance in its category, significantly outperforming baselines, achieving up to a 39\% gain in the final LLM-as-a-Judge score (2.17 to 3.02) and a 117\% boost in ID consistency (1.75 to 3.80). Human evaluators also prefer our method in up to 62\% of cases.
\end{itemize}

\section{Related Work}

\subsection{Video Summarization and Vision-Language Models}
Video summarization has evolved from early unsupervised approaches~\cite{Mahasseni_2017_CVPR,ZANG202326} to supervised methods benchmarked on datasets like TVSum~\cite{TVSum,ZHANG2024123568,BANJAR2024109795}. However, these models often struggle to capture the global narrative arc of long-form content. 
The advent of Vision-Language Models (VLMs), such as CLIP~\cite{radford2021learning} and Video-LLaMA~\cite{zhang2023video,V2Xum-LLM}, has revolutionized multimodal understanding. 
Nevertheless, applying general-purpose VLMs to full-length movies faces critical hurdles. 
One major limitation is the lack of factual grounding, where models frequently fail to track specific character identities, resulting in hallucinated descriptions~\cite{ji2023survey}. 
Another significant bottleneck is the computational constraint imposed by the quadratic complexity ($O(n^2)$) of self-attention, which makes processing hour-long videos in a single pass prohibitive.

To address the long-sequence challenge, efficient attention mechanisms like Longformer~\cite{beltagy2020longformer} and Ulysses~\cite{jacobs2024system} have been proposed to reduce complexity, yet they do not explicitly model narrative structure. 
An alternative strategy, progressive abstraction~\cite{StoryTellerIL}, involves iteratively condensing fine-grained segments into higher-level representations. 
This approach is particularly well-suited for movies as it mirrors the inherent cinematic hierarchy from shots to scenes and acts.

\subsection{Tool-Augmented Language Models} \label{section2.3}

Augmenting LLMs with external tools has emerged as a promising direction to mitigate their intrinsic limitations. Seminal work on Toolformer \cite{schick2023toolformer} demonstrated that language models can learn to autonomously execute API calls. This paradigm has been further advanced by frameworks like AgentThink \cite{qian2025agentthink}, which integrates Chain-of-Thought (CoT) reasoning for autonomous driving, and ChatHuman \cite{ChatHuman}, which enables dynamic tool selection in response to user inputs. This ``Tool-Augmented'' approach allows generalist models to leverage specialist capabilities and external knowledge bases, achieving performance beyond their standalone capacity.

\begin{figure*}[t]
\centering
\includegraphics[width=0.9\textwidth]{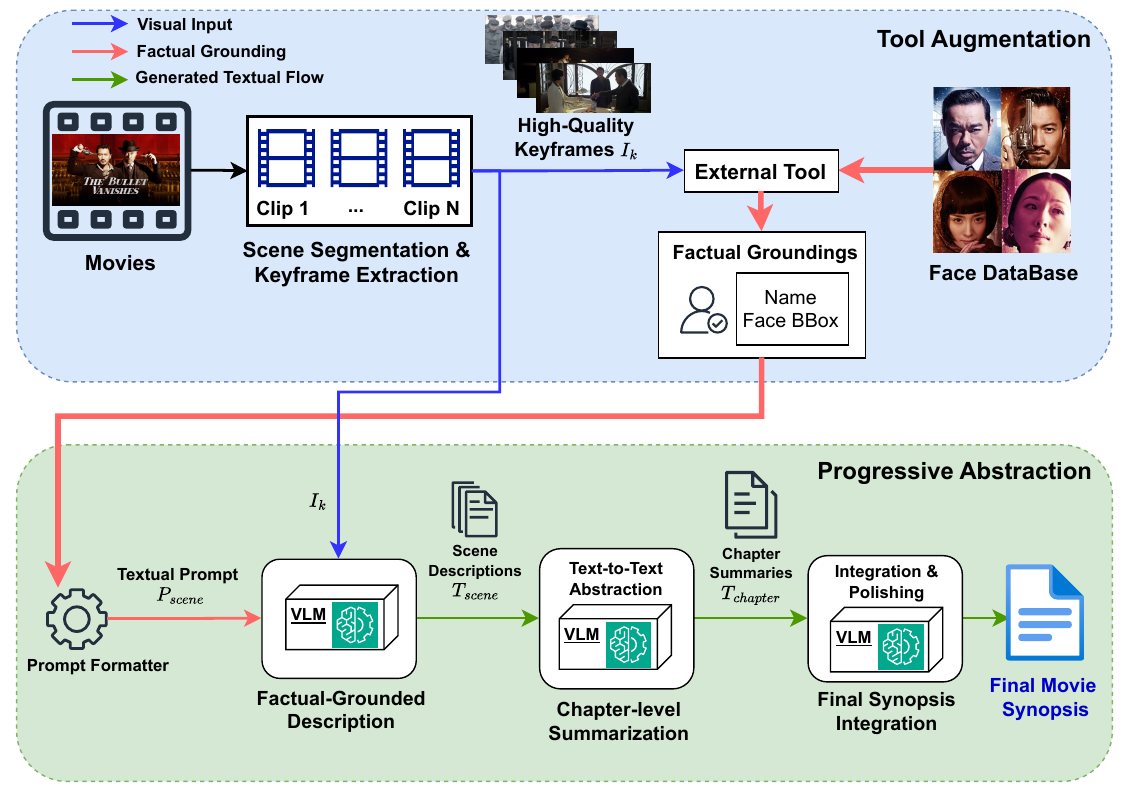}
\caption{The overall architecture of our proposed \textbf{MovieTeller} framework. The framework initiates by processing a long-form video to extract high-quality keyframes through scene segmentation and a quality gate. A key innovation is the subsequent tool-augmented stage, where an expert tool provides factual groundings (character ID, BBox) to a VLM, ensuring ID-consistent scene descriptions. This information is then progressively abstracted, first into chapter summaries and finally integrated into the complete movie synopsis.} 
\label{architecture}
\end{figure*}

\section{Method}

\subsection{Problem Definition}
Given a long-form video input $\mathcal{V}$ and its associated cast metadata $\mathcal{M} = { (n_i, p_i) }$ with character names $n_i$ and the corresponding actor images $p_i$, the system $\Gamma$ should generate a structured, detailed and accurate video description $\mathcal{D}$ that incorporates grounded character names.
\begin{equation}
    \mathcal{D} = \Gamma(\mathcal{V};\; \mathcal{M})
\end{equation}
To generate a high-fidelity, factually accurate, and narratively coherent synopsis from a full-length movie, we propose \textbf{MovieTeller}, a framework based on tool-augmentation and progressive abstraction. The core design philosophy is to decompose the complex, end-to-end summarization task into a pipeline of more manageable and specialized sub-tasks. This modular, training-free approach allows us to leverage the strengths of different state-of-the-art expert models at appropriate stages, ensuring both high performance and adaptability. As illustrated in \textbf{Figure \ref{architecture}}, our framework consists of three key stages: (1) Scene Segmentation and Keyframe Extraction; (2) Factual-Grounded Scene Description Generation via Tool Augmentation; (3) Progressive Abstraction.

\subsection{Scene Segmentation and Keyframe Extraction}

The initial step in our pipeline transforms the raw video stream into a concise set of representative keyframes. To achieve this, we first employ PySceneDetect\cite{ref_pyscenedetect} to partition the video $\mathcal{V}$ into a sequence of semantically coherent scenes $\mathcal{S}$, based on significant changes in visual content, as formalized in Eq. (\ref{seg}).

\vspace{-5pt}

\begin{equation}
\mathcal{S} = {(s_{1}, e_{1}), (s_{2}, e_{2}), \dots , (s_{n} , e_{n} )} = \mathcal{F}_{scene}(\mathcal{V})
\label{seg}
\end{equation}

\noindent where $\mathcal{F}_{scene}$ 
  represents the segmentation function and $(s_{i}, e_{i})$ denotes the start and end frames of the $i$-th scene. From each resulting scene, we then extract a single keyframe. To avoid selecting non-informative transitional frames (e.g., black screens), we implement a crucial Keyframe Quality Gate. This gate validates each candidate frame through a dual-threshold check on its mean brightness and pixel standard deviation, ensuring that only visually rich content is passed to the VLM for analysis.

\vspace{-5pt}

\subsection{Factual-Grounded Scene Description Generation}

This stage operationalizes our core contribution: augmenting a general-purpose VLM with a specialist tool to ensure factual accuracy, so as to improve the accuracy and consistency of results. 
Although state-of-the-art VLMs effectively extract visual elements such as human actions and interpersonal relationships, they struggle with maintaining character identity consistency. 

To tackle this issue, we leverage publicly available cast metadata to pre-construct a face database $\mathcal{B}$, containing the names $id$ and pre-computed embedding vectors of the main characters $e_{id}$. For each face $f$ detected in a keyframe with embedding $e_f$, its identity is determined by finding the closest match in the database:

\vspace{-10pt}

\begin{equation}
\text{ID}(f) = \underset{id \in \mathcal{B}}{\arg\max} \left( \frac{e_f \cdot e_{id}}{\|e_f\| \|e_{id}\|} \right)
\label{eq:face_recognition}
\end{equation}

Association is performed only when the similarity score exceeds a predefined confidence threshold $\tau_{\text{id}}$, therefore obtaining % The output of this tool-processing step is 
a set of Factual Groundings, denoted $G = \{(n_j, b_j)\}$, which combines identity $n_j$ and bounding box $b_j$. Based on $G$, we annotate each frame % with the face bounding box and the corresponding id 
to obtain the refined keyframes $F^{'} = \{f_1^{'}, f_2^{'}, ... , f_n^{'}\} $.

Thus, guided by a carefully engineered prompt $\mathcal{P}_{\text{desp}}$ (Listing \ref{lst:scene_prompt}),
%(\textbf{Appendix \ref{vlm_prompt}})
VLM generates persona-grounded descriptions with accurate character identities $ C_{scene}$. 
By reframing VLM's task from open-ended to fact-constrained narration, our prompting technique enhances the reliability and coherence of outputs.

\vspace{-10pt}

\begin{equation}
    C_{scene}=\{c_1,c_2,...,c_n\} = \text{VLM}(F^{'}; \mathcal{P}_{\text{desp}})
\end{equation}

\begin{figure}[t!]
\begin{lstlisting}[basicstyle=\ttfamily\scriptsize,
                  breaklines=true,
                  frame=single,
                  breakatwhitespace=true,
                  captionpos=b,
                  caption={Prompt for frame description generation.},
                  label={lst:scene_prompt},
                  numbers=left,
                  numberstyle=\tiny\color{gray},
                  showstringspaces=false,
                  commentstyle=\color{green}
                 ]
You are a professional movie analyst. Please generate a summary for the movie scene represented by the following keyframe.
To assist you, I have identified the following characters and their locations in the image:
- The actor `{Character_Name_1}' is located within the bounding box {BBox_1}.
- The actor `{Character_Name_2}' is located within the bounding box {BBox_2}.
...

Please generate the scene summary by combining the character identities with the visual content:
\end{lstlisting}
\vspace{-15pt}
\end{figure}

\subsection{Progressive Abstraction}

To overcome the context length limitations of VLMs and construct a coherent narrative for the entire movie, we employ a two-stage abstraction process.

\noindent \textbf{Chapter-Level Summarization.} 
The scene descriptions generated in the previous step $C_{scene}$, are first grouped into sequential chunks, each forming a ``chapter". We then, again in parallel, instruct the VLM to summarize each chapter with $\mathcal{P}_{\text{sum}}$, 
%(\textbf{Appendix \ref{vlm_prompt}})
distilling the core plot progression, character motivations, and key turning points within that specific chapter, while continuing to use the identified character names.

\vspace{-10pt}
\begin{equation}
    c_k{'} = \text{VLM}(c_{k}; \mathcal{P}_{\text{sum}}),\ c_k\in C_{scene}, \, k = 1,2,...n
\end{equation}

\noindent \textbf{Final Synopsis Integration.}

In the final stage, all chapter-level summaries are concatenated into a unified draft and processed by the VLM with prompt $\mathcal{P}_{\text{synopsis}}$. This prompt instructs the model to adopt a screenwriter persona, synthesizing the inputs into a cohesive movie synopsis that captures the global narrative arc from exposition to resolution. This hierarchical approach mirrors human cognitive summarization patterns, ensuring the final output maintains both logical coherence and semantic depth.

\begin{equation}
    \mathcal{D}_{\text{final}} = \text{VLM}\left( \bigoplus_{k=1}^n c_k^{'} ; \mathcal{P}_{\text{synopsis}} \right)
\end{equation}

\section{Experiments}

In this section, we conduct a series of experiments to empirically validate the effectiveness of our proposed MovieTeller framework. Our evaluation is designed to answer three key research questions:
\begin{enumerate}
    \item Does our full framework outperform baseline methods in generating high-quality movie synopses?
    \item What is the specific contribution of providing factual groundings (both names and bounding boxes) to the VLM?
    \item How do the generated synopses compare qualitatively?
\end{enumerate}

\subsection{Experimental Setup}
\subsubsection{Dataset}
\begin{figure*}[t]
    \centering
    \vspace{-10pt}
    \includegraphics[width=0.24\textwidth]{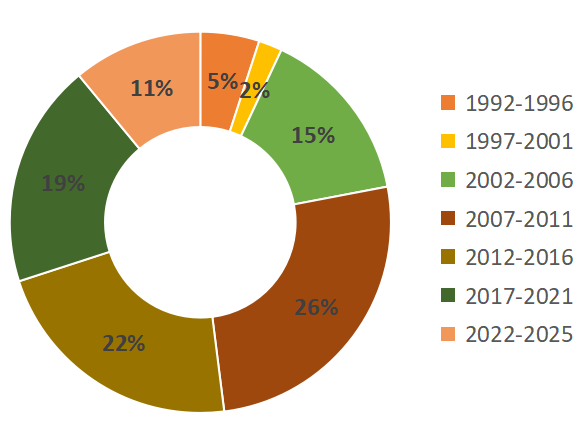} \hspace{3em} %\hfill
    \includegraphics[width=0.24\textwidth]{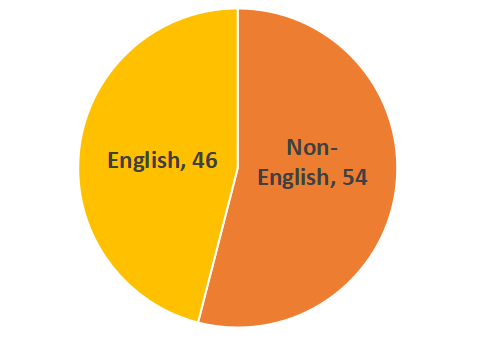} \hspace{3em} %\hfill
    \includegraphics[width=0.24\textwidth]{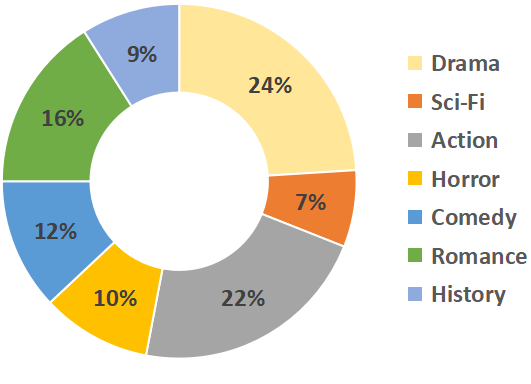}
    
    \vspace{-5pt}
    \caption{Dataset diversity statistics (100 movies). From left to right: distributions of release year, language, and genre.}
    \label{fig:dataset}
    \vspace{-10pt}
\end{figure*}

Our experiments were conducted on a dataset exceeds 10k minutes with 100 full-length movies, with each movie scoring above 5.0 on IMDb. 
\textbf{Figure \ref{fig:dataset}} depicts the distribution of the dataset, which contains at least five movies per genre across seven major genres and spans over three decades of release dates. The selection includes critically acclaimed titles with complex narratives, such as \textit{Farewell My Concubine}, \textit{Eat Drink Man Woman}, \textit{The Chronicles of Narnia}, 
and \textit{Iron Man 3}. This diversity mitigates potential bias towards specific genres or cinematic styles. For each film, we pre-build the required face database by collecting only 2 high-quality images of every main actor: one film still and one out-of-scene publicity photograph. 

\subsubsection{Implementation Details}
Our framework is implemented in Python using PyTorch. All experiments were run on a server equipped with 4x NVIDIA A40 GPUs (48GB VRAM each). The parallelized stages (scene description and chapter summarization) utilized all 4 GPUs. Our primary experiments utilize the Qwen2.5-VL-7B-Instruct \cite{bai2025qwen25vltechnicalreport} model as the core VLM. To demonstrate the versatility and generalizability of our framework, we also conduct ablation studies by replacing the core VLM with other state-of-the-art models, including InternVL3-8B \cite{zhu2025internvl3} and WeThink-Qwen2.5VL-7B \cite{yang2025wethink}. The face recognition ``tool" is powered by InsightFace \cite{Deng_2021_ICCV} with an ArcFace loss-based recognition head \cite{deng2018arcface}. For scene segmentation, we employ PySceneDetect \cite{ref_pyscenedetect} with the ContentDetector.

\subsubsection{Baselines and Ablation Models}

To comprehensively evaluate our approach, we compare our full MovieTeller framework against two ablation variants:
\begin{itemize}
\item \textbf{No-Hint (Baseline):} This is the most basic approach. The VLM generates scene descriptions based solely on the visual information from the keyframe, without any textual hints about character identities.
\item \textbf{Name-Only Hint (Ablation):} In this setting, we provide the VLM with only the names of the characters identified in the keyframe, but crucially, without their bounding box coordinates. This allows us to isolate and measure the specific impact of providing spatial grounding. % (the BBox).
\end{itemize}

\subsection{Evaluation Metrics}

Evaluating the quality of generated narratives, particularly abstractive summaries like movie synopses, is a notoriously challenging open problem. Unlike tasks with a single correct answer, there is no single ``ground truth" synopsis for a film. The plots of a movie plot can be validly summarized in numerous ways, each with different levels of detail and stylistic choices. This one-to-many mapping problem renders traditional n-gram-based metrics such as ROUGE \cite{lin2004rouge} and BLEU \cite{papineni2002bleu}, which measure lexical overlap against a single reference text, largely inadequate.
Therefore, to provide a more holistic and meaningful assessment, we employ a triple-pronged evaluation strategy that combines:

\subsubsection{Automatic Metrics}
As traditional n-gram-based metrics like ROUGE and BLEU are known to correlate poorly with human judgment for abstractive summarization, we report BERTScore \cite{zhang2019bertscore} for its ability to measure semantic similarity. We compute the F1 score between the generated synopsis and a reference synopsis obtained from a public encyclopedia. % (\textbf{Table \ref{tab:quantitative_results}}).

\subsubsection{LLM-as-a-Judge \cite{zheng2023judging}}

We employ Gemini 2.5 Flash \cite{comanici2025gemini} as an automated evaluator to assess the quality of final synopses. Leveraging its internal knowledge of canonical movie plots, the model scores each synopsis given the film's title. The evaluation covers four key dimensions: Factual Faithfulness,  ID Consistency \& Completeness, Narrative Coherence \& Flow, and  Conciseness \& Essence Capture. Each dimension is rated on a scale of 1 to 5, and the final metric is reported as the average of these component scores.

\subsubsection{Human Evaluation}

We further validated our approach through a human evaluation of 50 randomly sampled summaries. Evaluators performed a pairwise comparison based on perceived quality, offering a direct measure of practical utility that complements the detailed automated metrics.

\begin{table}[t!]
\centering
\caption{BERTScore(F1) comparison under three experimental settings. The best results are shown in bold.}
\label{tab:quantitative_results}
\resizebox{\columnwidth}{!}{
    \begin{tabular}{c|c|c|c}
    \hline
    \textbf{Method} & \textbf{Qwen2.5-VL} & \textbf{InternVL3} & \textbf{WeThink} \\ \hline
    \makecell{No-Hint}
    %No-Hint (Baseline)        
    & 0.612 & 0.616 & 0.618 \\ 
    \makecell{Name-Only}
     & 0.616 & 0.617 & 0.624 \\  
    \textbf{\makecell{MovieTeller (Ours)}} & \textbf{0.638} & \textbf{0.631} & \textbf{0.639} \\ 
    \hline
    \end{tabular}
}
\end{table}

\begin{table}[t!]
\centering
\small
\setlength{\tabcolsep}{5pt}
\caption{LLM-as-a-Judge evaluation results (1-5 scale). Scores are averaged across all test movies. The best results are shown in bold.}
\label{tab:llm_judge_results}
\resizebox{\columnwidth}{!}{
    \begin{tabular}{c|cccc|c}
    \hline
    \textbf{Method} & \textbf{Faith.} & \textbf{ID Cons.} & \textbf{Coherence} & \textbf{Concise.} & \textbf{Final} \\
    \hline
    \multicolumn{6}{l}{\textit{Base Model: Qwen2.5-VL-7B-Instruct}} \\
    \hline
    \makecell{No-Hint} & 2.49 & 1.80 & 2.24 & 2.17 & 2.18 \\
    \makecell{Name-Only} & 3.15 & 3.55 & 2.35 & 2.41 & 2.87 \\
    \textbf{\makecell{MovieTeller (Ours)}} & \textbf{3.21} & \textbf{3.65} & \textbf{2.52} & \textbf{2.45} & \textbf{2.96} \\
    \midrule
    \multicolumn{6}{l}{\textit{Base Model: InternVL3-8B}} \\
    \hline
    \makecell{No-Hint} & 2.51 & 1.75 & 2.20 & 2.21 & 2.17 \\
    \makecell{Name-Only} & 3.22 & 3.67 & 2.26 & 2.38 & 2.88 \\
    \textbf{\makecell{MovieTeller (Ours)}} & \textbf{3.34} & \textbf{3.80} & \textbf{2.50} & \textbf{2.44} & \textbf{3.02} \\
    \midrule
    \multicolumn{6}{l}{\textit{Base Model: WeThink-Qwen2.5VL-7B}} \\
    \hline
    \makecell{No-Hint} & 2.45 & 1.82 & 2.26 & 2.25 & 2.20 \\
    \makecell{Name-Only} & 3.19 & 3.65 & 2.38 & 2.42 & 2.91 \\
    \textbf{\makecell{MovieTeller (Ours)}} & \textbf{3.30} & \textbf{3.76} & \textbf{2.55} & \textbf{2.51} & \textbf{3.03} \\
    \hline
    \end{tabular}
}
\end{table}

\begin{table}[t!]
\centering
\caption{Human evaluation results. The values represent the preference rate (win rate, \%) for each method in a 3-way forced-choice task. For each base VLM, we report the percentage of times each method was selected as the most preferred. The most preferred are shown in bold.}
\label{tab:human_eval_results}
\resizebox{\columnwidth}{!}{
    \begin{tabular}{c|c|c|c}
    \hline
    \textbf{Method} & \textbf{Qwen2.5-VL} & \textbf{InternVL3} & \textbf{WeThink} \\ 
    \hline
    \makecell{No-Hint}       & 6\%  & 4\%  & 6\%  \\ 
    \makecell{Name-Only} & 34\% & 42\% & 32\% \\  
    \textbf{\makecell{MovieTeller (Ours)}} & \textbf{60\%} & \textbf{54\%} & \textbf{62\%} \\ 
    \hline
    \end{tabular}
}
\end{table}

\subsection{Results and Analysis}
\subsubsection{Quantitative Results}
The automatic metrics (Table \ref{tab:quantitative_results}) show MovieTeller achieves the highest BERTScore (e.g., 0.638 vs. 0.612 for Qwen2.5-VL), indicating better semantic alignment. LLM-as-a-Judge scores (Table \ref{tab:llm_judge_results}) provide nuanced validation: MovieTeller achieves the highest final scores (e.g., 3.02 vs. 2.17 on InternVL3) and excels in ID Consistency, showing up to a 117\% relative gain (3.80 vs. 1.75). This improvement occurs because factual bounding boxes resolve visual ambiguities, effectively preventing character hallucination. Finally, human evaluation (Table \ref{tab:human_eval_results}) confirms its superiority, with evaluators preferring MovieTeller (up to 62\% win rate) over the baseline (6\%) and Name-Only ablation (32\%). The gap between the Name-Only model and our full framework demonstrates that merely supplying character names without spatial coordinates still leaves the VLM guessing ``who is who" in dense, dynamic frames, whereas our BBox grounding enforces strict, unambiguous visual-textual alignment.

\begin{figure} % [H]
    \centering
    \includegraphics[width=1\linewidth]{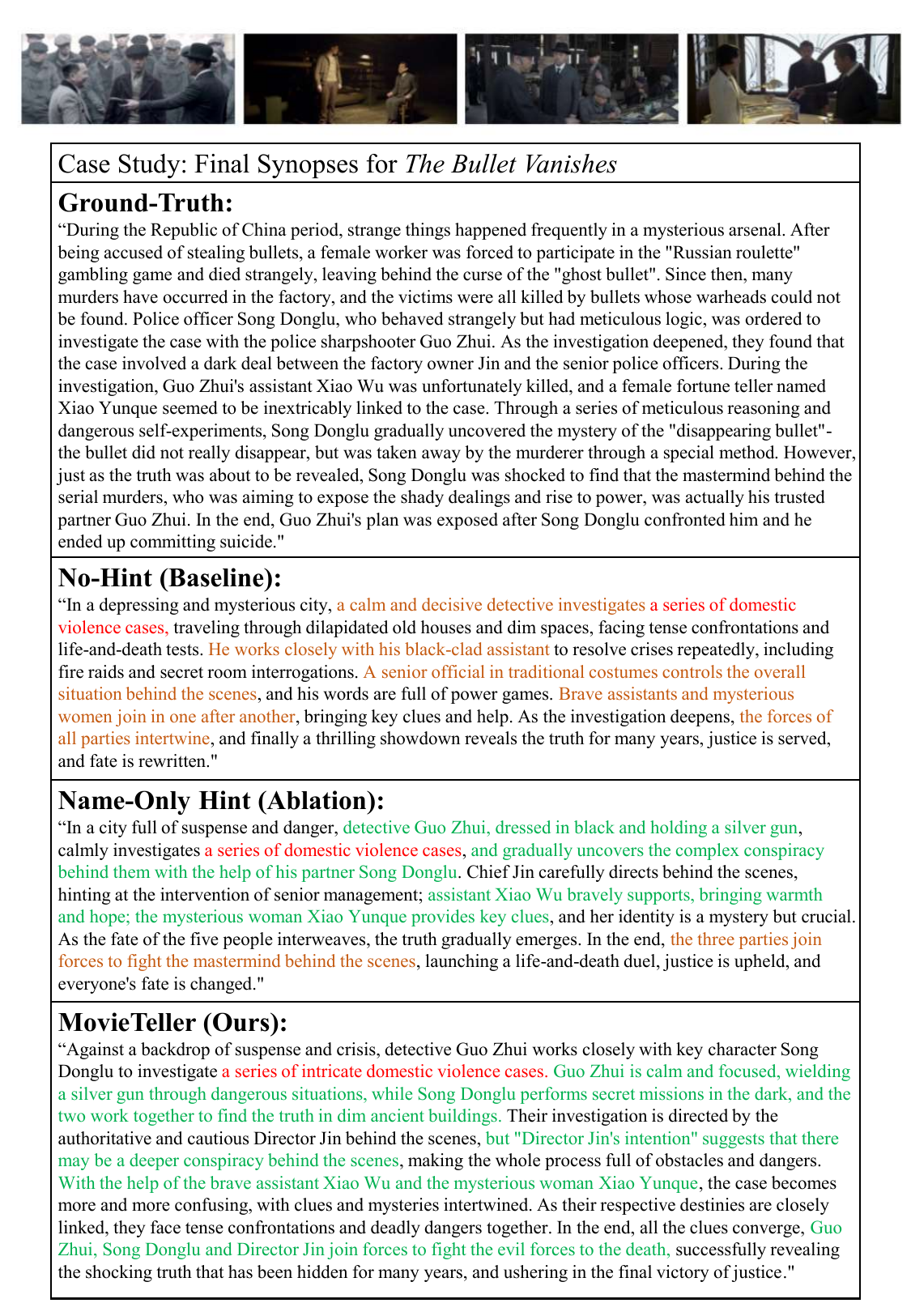}
    \caption{Qualitative comparison of the final, full-movie synopses generated by each method. Color legend: \textcolor{green!60!black}{Green} for correct and detailed information, \textcolor{orange!80!white}{Orange} for vague information, \textcolor{red}{Red} for incorrect information. MovieTeller produces a synopsis that is superior in both factual detail and narrative depth.}
    \label{fig:case_study}
\end{figure}

\subsubsection{Qualitative Analysis}

\textbf{Figure \ref{fig:case_study}} presents a qualitative comparison on the film \textit{The Bullet Vanishes}. The No-Hint baseline produces a generic summary that fails to identify characters, making the plot notoriously difficult to follow. While the Name-Only ablation improves identity recall, it misses the nuanced character interactions essential for driving the narrative. In contrast, MovieTeller, benefiting from factual grounding, generates a synopsis that is not only ID-consistent but also captures complex plot dynamics (e.g., the specific tension with ``Director Jin''), demonstrating superior narrative coherence over the baselines.

\section{Conclusion}

In this paper, we introduced \textbf{MovieTeller}, a novel framework to address critical challenges in long-form video summarization, namely the lack of ID-consistent character identification and fractured narrative coherence. MovieTeller leverages a training-free, tool-augmented progressive abstraction pipeline. By invoking a specialized face recognition model as an external tool to establish Factual Groundings, our method compels a general Vision-Language Model (VLM) to generate descriptions anchored to verifiable character identities. This, combined with our progressive abstraction mechanism, effectively mitigates long-context limitations and significantly enhances the factual accuracy, character consistency, and narrative quality of the final synopsis, as demonstrated by our experiments.
While promising, MovieTeller's performance is dependent on its upstream components, such as the completeness of the face database, and it currently omits the audio modality. These limitations present clear directions for future work. A primary focus will be the integration of audio, linking dialogue to visually identified characters via speaker diarization to create richer, dialogue-aware summaries. Furthermore, exploring dynamic tool-use protocols, where the VLM learns to invoke tools contextually, presents another exciting avenue. The modularity of MovieTeller also makes it highly adaptable for other domains, including sports analysis and documentary summarization.

% \section*{Acknowledgment}
% The preferred spelling of the word ``acknowledgment'' in America is without 
% an ``e'' after the ``g''. Avoid the stilted expression ``one of us (R. B. 
% G.) thanks $\ldots$''. Instead, try ``R. B. G. thanks$\ldots$''. Put sponsor 
% acknowledgments in the unnumbered footnote on the first page.

\bibliographystyle{IEEEtran}
\bibliography{mybibliography}

\end{document}